# Generative AI for Health Technology Assessment: Opportunities, Challenges, and Policy Considerations

Rachael L. Fleurence, PhD, MSc[1], Jiang Bian, PhD[2 3 4], Xiaoyan Wang, PhD[5 6], Hua Xu, PhD[7], Dalia Dawoud, PhD[8 9], Mitch Higashi, PhD[10], Jagpreet Chhatwal, PhD[11 12]

[1] Office of the Director, National Institute of Biomedical Imaging and Bioengineering, National Institutes of Health, Bethesda, MD, United States

[2] Health Outcomes and Biomedical Informatics, College of Medicine, University of Florida, Gainesville, FL, United [3] Biomedical Informatics, Clinical and Translational Science Institute, University of Florida, Gainesville, FL, United States

[4] Office of Data Science and Research Implementation, University of Florida Health, Gainesville, FL, United States

[5] Tulane University School of Public Health and Tropical Medicine, New Orleans, LA, United States

[6] Intelligent Medical Objects, Rosemont, IL, United States

[7] Department of Biomedical Informatics and Data Science, School of Medicine, Yale University, New Haven, CT, United States

[8] National Institute for Health and Care Excellence, London, United Kingdom

[9] Cairo University, Faculty of Pharmacy, Cairo, Egypt

[10] ISPOR - The Professional Society for Health Economics and Outcomes Research, Lawrenceville, NJ, United States

[11] Institute for Technology Assessment, Massachusetts General Hospital, Harvard Medical School, Boston, MA, United States

[12] Center for Health Decision Science, Harvard University, Boston, MA, United States

**Funding:** Dr Dalia Dawoud reports partial funding from the European Union's Horizon 2020 research and innovation programme under Grant Agreement No. 82516 (Next Generation Health Technology Assessment (HTx) project. No other funding was received.

**Acknowledgements**: The authors thank Dr Tala Fakhouri for her comments on earlier versions of this manuscript.

arXiv:2407.11054 [cs.LG] Sept 21, 2024 Version 3

**Generative AI for Health Technology Assessment: Opportunities, Challenges, and Policy Considerations**

**Abstract**

**Objective:** To provide an introduction to the uses of generative Artificial Intelligence (AI) and foundation models, including large language models (LLMs), in the field of health technology assessment (HTA).

**Methods**: We reviewed applications of generative AI in three areas: systematic literature reviews, real world evidence (RWE) and health economic modeling.

**Results**: (1) Literature reviews: generative AI has the potential to assist in automating aspects of systematic literature reviews by proposing search terms, screening abstracts, extracting data and generating code for meta-analyses; (2) Real World Evidence (RWE): generative AI can facilitate automating processes and analyze large collections of real-world data (RWD) including unstructured clinical notes and imaging; (3) Health economic modeling: generative AI can aid in the development of health economic models, from conceptualization to validation. Limitations in the use of foundation models and LLMs include challenges surrounding their scientific rigor and reliability, the potential for bias, implications for equity, as well as nontrivial concerns regarding adherence to regulatory and ethical standards, particularly in terms of data privacy and security. Additionally, we survey the current policy landscape and provide suggestions for HTA agencies on responsibly integrating generative AI into their workflows, emphasizing the importance of human oversight and the fast-evolving nature of these tools.

**Conclusions:** While generative AI technology holds promise with respect to HTA applications, it is still undergoing rapid developments and improvements. Continued careful evaluation of their applications to HTA is required. Both developers and users of research incorporating these tools, should familiarize themselves with their current capabilities and limitations.

**Highlights**

1. Generative AI and foundation models, including large language models (LLMs), have the potential to transform health technology assessment (HTA) by automating elements of evidence synthesis, enhancing real-world evidence generation, and streamlining health economic modeling.

2. Despite their potential, current generative AI applications are in their early stages and present limitations, including issues of scientific validity and reliability, risk of bias and impact on equity, and regulatory and ethical considerations. These will continue to require careful evaluation and human oversight in their application to HTA for the foreseeable future.

3. The article reviews the current policy landscape of generative AI and the authors offer suggestions for HTA agencies, emphasizing the need for integrating the use of foundation models into guidance, harmonizing standards, and investing in training to responsibly integrate generative AI into their assessment processes.

**Key Words**

Artificial Intelligence, Generative AI, Large Language Models, Systematic Reviews, Economic Modeling Methods, Real World Evidence

**Generative AI for Health Technology Assessment: Opportunities, Challenges, and Policy Considerations**

## Introduction

Artificial intelligence (AI) has been defined as "the science and engineering of making intelligent machines" [1]. Advancements in AI and especially machine learning (ML) have been rapidly progressing for several decades [2,3]. However, the recent emergence of generative AI technologies such as large language models (LLMs) has ushered in a transformative era across numerous fields, including science, medicine, and other areas of human activity [4-6] (**Box 1**) (**Figure 1**).

Figure 1: Relationship between AI, Gen AI, Foundation Models and LLMs

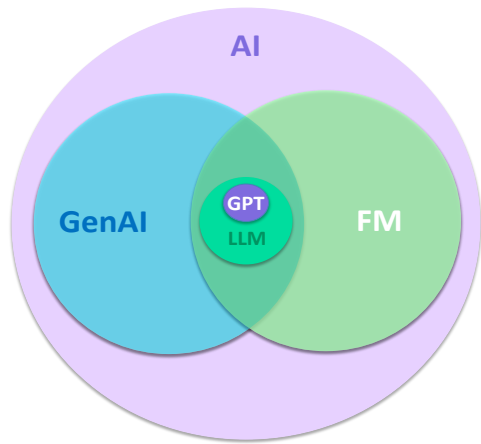

AI=Artificial Intelligence; FM= foundation model; Gen AI = generative artificial intelligence;
GPT= Generative Pre-Trained Transformer; LLM= large language model.

Generative AI employs sophisticated ML models, particularly a class of deep neural networks, that can generate language, images, and code in response to free text prompts provided by users. These models are trained using self-supervised learning techniques on vast amounts of existing data, allowing them to identify patterns and relationships in the underlying data autonomously. Their training requires substantial computational power, often relying on specialized hardware, especially graphics processing units (GPUs) [7].

In the field of Health Economics and Outcomes Research (HEOR), generative AI has the potential to transform evidence generation methods used in health technology assessments (HTA). Despite this promise, the exploration of opportunities, limitations, and risks associated with developing, evaluating, and deploying these models for supporting different approaches to evidence generation remains sparse, given the recency of these new tools. This article seeks to fill this gap by offering a review of the landscape of generative AI's applications in methodological areas relevant to HTA development.

The perspective taken in this article is that, in this current phase, foundation models should be used to augment human activities for which humans remain fully accountable. Researchers are ultimately responsible for the quality and accuracy of results generated with the help of foundation models and for using existing guidelines and checklists to report results. Nevertheless, foundation models do add an additional layer of complexity that needs to be acknowledged and addressed during the analysis and the reporting.

The immediate audience of the article is HEOR professionals, but we anticipate that it will also be useful to other stakeholders, particularly HTA evidence developers and users, regulators, and manufacturers. The article is structured as follows. We begin with a concise history of AI and overview of generative AI. Next, we examine the opportunities and challenges of utilizing generative AI to support significant aspects of HEOR. This includes (1) literature reviews and evidence synthesis, (2) Real-World Evidence (RWE), (3) health economic modelling. We continue with a description of the current limitations and risks associated with generative AI tools. Next, we provide a summary overview of the current policy landscape on AI. We conclude with the results of a brief survey of HTA agencies gauging their current use of tools with generative AI and provide suggestions to HTA agencies on preparing for a more widespread adoption of generative AI tools.

## A brief history of AI and generative AI

AI has evolved significantly since its conception, tracing a path of remarkable technological advancements. Introduced by Alan Turing in 1950, the Turing Test challenged the ability to distinguish between responses from a machine and a human during a natural language conversation, setting a foundational criterion for machine intelligence [8]. In the following decade, significant progress was made with the development of symbolic methods to process natural language, such as the ELIZA chatbot in the 1960s [9]. In the medical field during the 1970s, expert systems like Stanford University's MYCIN demonstrated capabilities in reasoning under uncertainty and providing decision rationales, using rule-based AI [10].

The 1990s ushered in the era of ML algorithms with the development of techniques such as Support Vector Machines (SVM) and Random Forests. These methods excelled in tasks across various domains by effectively handling pattern recognition and decision-making processes. This period marked a shift from rule-based to data-driven AI, setting the stage for the next big leap. The introduction of deep learning in the 2000s further revolutionized AI capabilities [11]. Deep learning models based on neural networks, enabled a wide range of complex applications from image recognition to natural language processing (NLP). It also led to breakthroughs such as IBM Watson and AlphaGo—systems that famously outperformed humans in Jeopardy and Go, respectively [12,13]. While IBM Watson was not eventually successful in healthcare applications, AlphaGo led to the development of AlphaFold, which revolutionized the field of structural biology by solving the protein folding problem [14,15].

The emergence of generative AI technologies, especially the LLMs from OpenAI's GPT series, and the debut of ChatGPT in November 2022, marked a paradigm shift in the field of AI. Generative AI focuses on creating new content (e.g., text, images, videos) by learning from large existing data and it differs substantially from earlier AI technologies in several key aspects. First, LLMs employ the innovative transformer architecture, which features self-attention mechanisms to evaluate the relevance of each word in a sentence. This enhances the model's grasp of context and linguistic subtleties [16]. Pre-trained on vast text corpora, LLMs utilize billions of parameters to excel across a variety of tasks, such as summarization, translation, question answering, and code generation. These models are often called "foundation models" as the fundamental basis for a wide variety of downstream tasks. Since the development of ChatGPT, numerous LLMs have been created, consistently setting new benchmarks in performance across many tasks [7]. Additionally, these generative AI technologies offer user-friendly interfaces that allow for direct interaction through natural language text or speech, significantly reducing the need for structured data preparation typical of earlier ML models like SVMs. LLMs can also quickly adapt to new tasks through natural language instructions and in-context learning.

In summary, generative AI technologies like LLMs represent a fundamental shift in how machines can understand and interact with human language, offering a breadth of capabilities and flexibility that were not possible with earlier AI models. Many applications of LLMs to different data modalities (e.g., text, images, and videos) have been investigated across different industrial sectors including healthcare and life science [17,18].

**Applications of generative AI in literature reviews and evidence synthesis**

This section identifies applications of generative AI, including the use of foundation models such as LLMs in systematic literature reviews (SLRs) and meta-analyses. At this phase of development, foundation models should be understood as ways to augment human tasks and not as autonomous replacements for humans. There are several aspects of SLRs that can be augmented with foundation models. LLMs can assist in generating a search strategy by proposing Mesh terms and keywords to input in biomedical search engines such as PubMed [19]. Nevertheless, "hallucinations" which are instances where the output generated is factually incorrect, misleading or fabricated, can occur, and since LLM-generated citations might be fabricated, manually verifying the terms proposed by the models is required [19,20]. Search engines are also improving their search returns with the integration of AI supported tools and the use of retrieval augmented generation (RAG), a retrieval system to find relevant information to produce more contextually accurate and factual outputs, to anchor the model's response to verifiable references [21]. Researchers have evaluated the capacities of foundation models to automate abstract and full text screening. When screening full text literature using highly reliable prompts, which are precise inputs guiding the LLM to produce accurate and relevant responses, GPT-4 has been shown to have comparable performance to humans [22]. For example, Guo et al. assessed the performance of OpenAI GPT (GPT-3.5 Turbo and GPT-4) in automating screening and found a high level of agreement between the model-generated results and a human reviewer [23]. They also found that the model was able to explain its reasoning for excluding papers and correct its initial decisions [23]. Robinson et al. have also evaluated LLMs' capacity to provide exclusion reasoning for abstracts and full texts [24]. Hasan et al compared the agreement between GPT-4 and human reviewers in assessing the risk of bias using the Cochrane Collaboration's risk of bias tool. Their case study demonstrated a fair agreement between the LLM and the human reviewer [25]. Several studies have tested the data extraction capacities of LLMs. Reason et al. found that GPT-4 >99% accuracy in replicating the data extraction in 4 network meta-analyses [26]. Gartlehner et al. assessed the performance of CLAUDE-2 in extracting data elements from published studies compared to human extraction [27] and found that CLAUDE-2 achieved an overall accuracy of 96.3%. This study also provides examples of prompt engineering. Schopow et al. found a high degree of concordance between ChatGPT (GPT-3.5 and 4.0) and human researchers on the data extraction task [28]. Nevertheless, using GPT-4, another study only found moderate performance in assisting with data abstraction [22]. In addition to data extraction, the foundation model can also be directed to generate the code to conduct meta-analyses. Reason et al. found that the LLM used was able to generate error-free code in R for network meta-analyses [26]. However, Qureshi et al. found errors in code generated in an earlier GPT model, GPT-3.5 [19]. It should be noted that the presence of errors may be partly attributed to earlier versions of GPT, but that other factors may also contribute to this such as such as ambiguous or incomplete prompts and the inherent complexity of the coding tasks. Finally, report writing has also been evaluated with researchers finding that the LLM is able to generate reasonable drafts [26].

In summary, there is promise in using foundation models to support a range of tasks required in SLRs but the use of LLMs to augment human work in systematic literature reviews (SLRs) presents several limitations. These include potential inaccuracies in the outputs, such as errors in classifying abstract or in extracting data, as well as the generation of fabricated content, such as non-existent citations. While some studies have demonstrated that LLMs can achieve reasonable accuracy compared to human-performed tasks, this is not universally true, underscoring the necessity for continuous human oversight and validation [20]. Other limitations include difficulties in replicating results across different LLMs, which can be due to different user inputs or to the systems themselves. Currently Gen AI tools should be used to augment human tasks not autonomously replace them.

**Applications of generative AI to real-world evidence (RWE):**

Increasingly, RWE is being used by a wide range of stakeholders in the HTA process [29]. The potential benefits of using generative AI in RWE include increased efficiency in data processing and analysis and enhanced accuracy and consistency by minimizing human errors and standardizing the evidence generation processes. Unstructured notes in EHRs, such as radiology reports and physician notes, are important for RWE generation [30] and the use of LLMs can make information extraction from notes easier and more efficient [31]. A technique known as "few-shot learning", where a model is given a few examples in the prompt to guide its response, has been shown to be useful to extract variables from unstructured notes [32]. Domain-specific LLMs like GatorTron, GatorTronGPT and Me LLaMA, which are trained using large clinical texts have been shown to improve the accuracy of outputs [33-35].   Additionally, generative AI is increasingly being customized for images and other formats of real-world data to generate evidence [36]. There are still limitations when applying LLMs to the area of RWE. These include the generation of inaccuracies and fabrications. For example, one study found that a range of foundation models had less than 50% accuracy in mapping descriptive text to the correct ICD and CPT codes [37]. Potential for bias and privacy risk will further be discussed in the section on limitations.

**Applications of generative AI to health economic modeling**

Generative AI has the potential to support different phases of economic model development, including model conceptualization, parameterization, model implementation, and evaluation and validation of model results [38-40].   Chhatwal et al. explored the use of foundation models—GPT-4 and Bing Chat—in the conceptualization of a Markov model for hepatitis C and parameterization of the model [41]. They found significant variability in how these models conceptualized disease progression. The study emphasized the significance of expert guidance in utilizing LLMs for HEOR model development to ensure accuracy and reliability of the results. .

Reason et al. demonstrated the capability of GPT-4 to recreate published three-state partition survival models for non-small cell lung cancer and renal cell carcinoma [42]. This study provides an in-depth use of prompt engineering, including chain-of-thought prompting, to improve results. While the model construction was replicated with high accuracy, this approach required human intervention to provide contextual information on top of information specifying the model assumptions, methods and parameter values. This suggests that although generative AI can create a model code, significant human expertise is still necessary to guide the model conceptualization and parameterization. In contrast, a study by Ayer et al. demonstrated the proof-of-concept of fully replicating a published 'simple' health economic model, including extraction of model structure and parameters, code development, and evaluation of results without human involvement.[43] However, further research is needed for fully replicating complex models. Generative AI could also play an important role in structural uncertainty analysis, a typically resource-intensive process [39].  By automating this aspect, generative AI can help in identifying and mitigating potential weaknesses in model structure more efficiently.

Automation of different stages of model development could lead to more efficient model development, reducing time and resource expenditure in the future, but current AI tools will augment and support human experts rather than replacing them entirely. Researchers should be aware of the limitations when using LLMs to in the development of health economic models. These include the potential for inaccuracies, such as erroneous parameters or coding errors, and the potential for fabricated data or parameters.

**Limitations of generative AI in HTA applications**

While generative AI offers promising applications in areas relevant to HTA, caution should be exercised when evaluating studies employing these technologies [4]. Several frameworks have been developed to help guide the responsible use of AI in health care, many of which remain relevant to generative AI [44-46]. Here, we briefly review and focus on three different categories of limitations of generative AI and foundation models: 1) Scientific validity and reliability, 2) Bias, equity and fairness and 3) Regulatory and ethical considerations.

*Scientific validity and reliability*

As outlined in the introduction, we recommend that Gen AI tools should currently be used to support human activities with humans remaining accountable. While researchers are responsible for results' accuracy and reporting, foundation models introduce a new level of complexity that must be carefully managed during analysis and reporting. In the following section, we review some features of foundations models that might impact scientific validity and reliability. Because LLMs are trained on large corpora of publicly available information from the web, errors may be introduced especially in domain specific areas like health [4], requiring careful review of results. A well noted limitation has been the generation of "hallucinations" [20,47]. This phenomenon occurs because models generate context based on learned statistical associations between words and sentences, not from an understanding of the knowledge in the corpora. Several strategies might be employed to improve the factual correctness of model outputs. First, improving the instructions used in the prompts provided to the models, a process known as prompt engineering, has been shown to improve the accuracy of the results. For example, chain-of-thought prompting is a technique in which a model is guided to generate step-by-step explanation or its reasoning process while arriving at an answer [48]. Few-shot learning, an in-context learning technology that prompts an LLM with several examples of a specific task, is another approach to improve model performance [49]. Second, the use of RAG allows the model to retrieve relevant information from external sources [21]. This has been shown reduce the incidence of hallucinations and factually incorrect output [50]. Third, foundation models can be fine-tuned with domain-specific data and knowledge, thus improving their performance on answering domain-specific questions [33,51,52].

In science findings should also be reproducible and foundation models add a level of complexity because they are more opaque than traditional statistical tools and learn patterns from data which can introduce variability in the generation of the model parameters. Model output variation can stem from either user factors (e.g., inexperience working with foundation models, suboptimal prompts) or inherent differences in the AI models. Researchers have proposed various approaches to support the reproducibility of studies and the accuracy of results. For example, Hasan et al. propose a framework for integrating foundation models into SLRs [25]. Reason et al. run their models over 20 times to compare the results from different runs [26]. Open sharing of data, code and results as well as the adoption of standards for reporting and transparency will likely help improve reproducibility [53,54]. Finally, transparency in this area is challenging because of the 'black box' nature of LLMs, meaning it is unclear how models generate answers from inputs. Strategies proposed to improve transparency have included requirements for the outputs to cite which part of the dataset contributed to the answer, and 'explainable' AI [4]. To achieve these goals and address the challenges above, the involvement of all stakeholders, with consideration of human-AI teaming, including human oversight, is critically needed in using these models [55].

*Bias, Equity and Fairness*

Foundation models can propagate or amplify biases introduced at many steps from model inception to deployment, resulting in systematic differences that may exacerbate inequities and cause harms to individuals and communities [56,57]. First, systemic bias (or institutional or historical bias) may occur in the use of historical data to train foundation models, where marginalized groups may be underrepresented in datasets because of barriers to access to resources, perpetuating existing systemic biases in society [58]. Second, computational and statistical biases may stem from errors that occur when the sample is not representative of the population and the algorithm produces a result that differs from the true underlying estimate [59,60]. Bias may also occur when specific population groups are excluded from data collection, training, testing or subsequent analyses [56].

Several strategies have been proposed to assess and mitigate the risk of bias in foundation models. Surveys on this topic have been published [61-63]. One review of methods for addressing AI fairness and bias in biomedicine identified distributional and algorithmic approaches for addressing bias [61]. Distributional approaches include strategies such as data augmentation, data perturbation and data reweighting, such as the generation of synthetic datasets to ensure balanced representation of all demographic groups [64]. Federated learning may offer some potential to address biases, where models are trained on multi-institutional data without the need for data sharing, with the idea that access to more diverse populations in many different health systems may mitigate biases inherent to localized datasets [56]. Algorithmic approaches change the algorithm itself and include approaches such as adversarial learning and loss-based methods, where the loss function is adjusted to penalize biased predictions [61].

*Regulatory and Ethical Considerations*

Regulatory frameworks focused on generative AI in biomedical research are under development in many countries. However, established data privacy and confidentiality laws such as the Health Insurance Portability and Accountability Act (HIPAA) in the United-States and General Data Protection Regulation (GDPR) in the European Union remain applicable. Gen AI models require a large amount of training data. It is important to avoid using data containing Protected Health Information (PHI) because absolute de-identification is not attainable, and re-identification risks remain non- zero [65,66].

Further Gen AI models have the ability to memorize the data that they were trained on, posing a risk of reproducing PHI that they memorized during training [67]. Using patient-level data or other sensitive, confidential information in commercial LLMs poses risks if data handling and compliance with privacy regulations are unclear, as the company providing the models may have access for further model training or other purposes. In contrast, with open-source models, the risk lies in the operator's control over data security, where improper handling could lead to leaks. In all cases, clear documentation and data governance must be in place when using public or proprietary datasets for AI training. Enhancements such as synthetic data that replicates real patient data without personal identifiers [68] and enabling computations on encrypted data [69], are being explored to augment privacy protections.

The issue of informed consent for AI driven research is becoming increasingly critical and is underscored by recent legal disputes over the use of patient EHR data to train LLMs [70]. Inaccurate information can lead to patient harm, for example if used to support an HTA submission that erroneously exclude underserved populations. In addition, emergent capabilities of LLMs may raise significant ethical challenges, particularly when the model's accurate performance cannot be transparently explained. For example, Giyocha et al. found that the AI model could identify a patient's self-reported race from imaging alone, across multiple imaging modalities, and often when clinical experts could not. The study was not able to isolate which features the models were using to make these accurate predictions. This raises important ethical issues that will need to be thoughtfully addressed because the recognition by AI models of race in medical images could produce different health outcomes for members of different racial groups [71]. In summary, given the early stages of the development and advances of generative AI, the themes discussed above will continue to be important and the object of warranted attention from many stakeholders within society.

**Policy landscape**

Policies regulating generative AI, including foundation models in healthcare, aim to ensure safe, ethical, and effective use of these technologies. The supplemental material includes a table that presents legal and policy documents related to the use of AI, generative AI, and foundation models from governments, regulatory bodies, and multi-stakeholder groups. The regulations address various aspects, such as risk management, accountability, transparency, and compliance with existing legal standards. The primary objective is to mitigate potential risks associated with AI, such as bias, privacy violations, and safety concerns, while fostering innovation and trust in AI systems.

Different regions adopt varied approaches: the EU implements a comprehensive, horizontal framework with stringent requirements for high-risk AI [72], while the UK, Japan, and the US favor sector-specific guidelines and flexible regulations [73,74],[75],[76,77].

International organizations have also issued guidance or statements on AI, including the Organization for Economic Co-operation and Development (OECD), the United Nations (UN) and United Nations Educational, Scientific and Cultural Organization (UNESCO) [78-80]. In the health care sector, the World Health Organization (WHO) has provided guidance to member states on deploying LMMs in health care, and outlining the risks such as the overestimating the benefits of using LLMs, propagating system-wide biases, negatively impacting the labor market, and increasing cybersecurity risks [81,82].

In the regulatory space, the European Medicines Agency (EMA) has issued a reflection paper on the use of AI/ML in the medicinal product life cycle, including product development, authorization, and post-authorization, emphasizing the need to meet all existing legal requirements, employ a risk-based approach, and promote AI trustworthiness [83]. The FDA has issued several discussion documents related to AI, specifically on the use of AI/ML-based software as a medical device [84,85] and more generally on AI use – not specific to generative AI – across the  drug development process, including drug discovery, nonclinical phase of drug development, the clinical phase including in clinical trials, post-marketing surveillance and in manufacturing [86-88].

In the field of HTA, we reviewed the websites of the main HTA bodies and did not identify statements or guidance on the use of AI in HTA submissions at the time of writing (May 2024). In August 2024, the National Institute for Health and Care Excellence issued a position statement on the use of AI in evidence generation[89]. The ISPOR HTA Roundtable conducted an informal survey with 18 members from North America, Latin America, Australasia and Europe, in April 2024 to gauge the level of internal use of generative AI in HTA agencies (They were not queried about the use of AI in submissions).  Several agencies mentioned that they were still in an exploratory phase with generative AI, seeking to better understand how it may support their work and be used by manufacturers in their submissions. Some groups were not using AI at this time and one agency had blocked the use of AI tools such as Chat-GPT at this time. Several agencies are evaluating software offered by private companies in the space of Systematic Literature Reviews (5 agencies) and one agency described experimenting with the use of generative AI in economic modelling.  Finally, some agencies mentioned internal use of AI tools for translation and for general education purposes. The learnings from these explorations of AI tools will likely inform the methods and processes guidance HTA agencies will need to develop to guide companies in using generative AI in HTA submissions [90].

We provide some suggestions for HTA agencies and policy makers for establishing procedures and processes to improve the use of generative AI in HTA submissions based on our experience in the HTA field (**Box 2**). We recommend establishing clear guidance on the appropriate use of LLMs, including examples of both suitable and unsuitable applications. Standardizing and harmonizing processes across agencies, ensuring transparent reporting will be important. A focus on health equity will enhance the quality and efficiency of submissions and ensure the appropriate populations are represented in HTA analyses. Additionally, investing in training for the HTA workforce will be essential for the responsible and inclusive use of generative AI technologies.

Multi-stakeholder groups are also working on frameworks for the responsible and trustworthy use of AI, including for example, the National Academies of Medicine, the US National Institute of Standards and Technology (NIST), the Coalition for Health AI and the European Commission's HTx H2020 (Next Generation Health Technology Assessment) project [44-46,60,90]. Finally, we note that scientific and medical journals are in the process of developing their policies with respect to the use of AI tools in publications submitted to their journals [91]. Of note, the new NEJM AI journal encourages the use of LLMs in submissions with the assumption that they will enhance the quality of research and support its democratization [92].

## Conclusion

The field of generative AI provides potentially transformative tools to augment and support the efficient generation of evidence to support HTA under human supervision. In this article, we have provided a brief overview of the development of generative AI and reviewed applications relevant to HTA including literature review and evidence synthesis, real world evidence, and health economic modeling. Despite their promise, it is important to acknowledge that these technologies, while rapidly improving, are still nascent and continued careful evaluation in their application to HTA is required. To ensure the responsible use and implementation of these tools, both developers and users of research must fully understand their limitations, including challenges related to scientific validity and reliability, risks of bias, potential impacts on equity, and critical regulatory and ethical considerations. Because of the nature of these models, there is added complexity in using them and evaluating the validity and reproducibility of their results. It is expected, as with all new technologies, that both user expertise with foundation models, and the actual performance of foundation models themselves, will improve rapidly in the near future. This is a transformative and exciting time in scientific and medical research, but scientific research is a complex endeavor that impacts human health and will continue to require careful validation and human oversight for the foreseeable future.

**Glossary**

- Artificial intelligence (AI): a broad field of computer science that aims to create intelligent machines capable of performing tasks typically requiring human intelligence.

- Chain of thought prompting: A technique where the model is guided to reason through a problem step-by-step in its response, by breaking down complex tasks into simpler parts to improve accuracy.

- Deep Learning: a subset of machine learning algorithms that uses multilayered neural networks, called deep neural networks. Those algorithms are the core behind the majority of advanced AI models.

- Few-Shot Learning: A method where a model is given a few examples in the prompt to guide its response, utilizing its pre-trained knowledge to produce accurate outputs with minimal data.

- Foundation model: a large scale pretrained models that serve a variety of purposes. These models are trained on broad data at scale and can adapt to a wide range of tasks and domains with further fine-tuning.

- Generative AI: AI systems capable of generating text, images, or other content based on input data, often creating new and original outputs.

- Generative Pre-trained Transformer (GPT): a specific series of LLMs created by OpenAI based on the Transformer architecture, which is particularly well-suited for generating human-like text.

- Hallucination: Incorrect output produced by a generative AI model that is not based on the input data or reality. This content is factually incorrect, misleading, or fabricated.

- Large Language Model: a specific type of foundation model trained on massive text data that can recognize, summarize, translate, predict, and generate text and other content based on knowledge gained from massive datasets.

- Machine learning (ML):  a field of study within AI that focuses on developing algorithms that can learn from data without being explicitly programmed.

- Multimodal AI: an AI model that simultaneously integrates diverse data formats provided as training and prompt inputs, including images, text, bio-signals, -omics data and more.

- Prompt:  the input given to an AI system, consisting of text or parameters that guide the AI to generate text, images, or other outputs in response.

- Prompt engineering: creating and adapting prompts (input) to instruct AI models to generate specific output.

- Retrieval Augmented Generation (RAG): A method in natural language processing that combines a generative model with a retrieval system to improve response accuracy. The retrieval system finds relevant information, which the generative model uses to produce more contextually accurate and factual outputs.

**Box 1. Key Definitions***

**Generative AI (Gen AI)**

AI systems capable of generating text, images, or other content based on input data, often creating new and original outputs.

- Example: DALL-E generating images from text descriptions.

**Foundation Models**

Large scale pretrained models that serve a variety of purposes. These models are trained on broad data at scale and can adapt to a wide range of tasks and domains with or without further fine-tuning. While LLMs are a subset of foundation models focusing specifically on language, foundation models can encompass other modalities, such as vision and multimodal tasks.

- Examples: GPT-4, Gemini, BERT, CLIP (Contrastive Language-Image Pretraining), LLama,

**Large Language Models (LLMs)**

A specific type of foundation model trained on massive text data that can recognize, summarize, translate, predict, and generate text and other content based on knowledge gained from massive datasets.

- Examples: GPT-4, Gemini, BERT, LLama, GatorTron.

*First draft of definitions developed using Chat-GPT4o, May 2024

**Box 2**. Suggestions to HTA Agencies and Policy Makers on the use of Large Language Models (LLMs)

- Develop guidance for assessing the appropriate use of LLMs for health technology assessment or endorse existing guidance.

- Provide examples of what constitutes acceptable use of LLMs in health technology assessments, including through use-cases and "non-use cases", i.e areas where it is not recommended to use LLMs in HTAs.

- Collaborate with other HTA agencies and regulatory agencies to harmonize standards, processes, and reporting requirements for the use of LLMs for health technology assessment.

- Develop guidance on and/or checklists for reporting the use of LLMs, including providing checks on validation, reproducibility, bias. These need not be stand-alone and might be incorporated into sections of existing methodological guidances.

- Assess whether LLM qualification or accreditation schemes, by vendor or dataset, would improve the efficiency and quality of submissions to HTA agencies.

- Engage in multi-stakeholder collaborations to support shared learning around the use of LLMs for HTA, develop best practices, and include the patient perspective.

- Ensure that health equity considerations are appropriately included in the guidance around the use of LLMs in HTA, such as ensuring that populations that have been historically marginalized are represented.

- Invest in training and preparing the HTA workforce and partners (e.g., academic collaborators, committee members) in the area of LLMs. Involve relevant experts where possible.

**Supplemental Table:** Laws, Policies and Guidance on the use of Generative AI from governments, regulatory bodies, and multi-stakeholder groups

| Entity | Availability of guidance on the use of AI and LLMs |
|---|---|
| **Governments** | |
| **Canada** | - The Artificial Intelligence and Data Act (AIDA) – Companion document (Canada, 2022) |
| **European Union** | - EU AI ACT (EU, 2024) |
| **G7** | - Hiroshima Process International Code of Conduct for Organizations Developing Advanced AI Systems (G7, 2023) |
| **Japan** | - Japan's Approach to AI Regulation and Its Impact on the 2023 G7 Presidency (Habuka, 2023) |
| **United-Kingdom** | - United-Kingdom, A pro-innovation approach to AI regulation: government response (UK Consultation, 2024)<br>- United-Kingdom, A pro-innovation approach to AI regulation (UK, 2023) |
| **United-States** | - Executive Order of the President on safe, secure and trustworthy AI (EOP, 2024)<br>- Blueprint for an AI Bill of Rights (EOP, 2022) |
| **International organizations** | |
| **Organization for Economic Cooperation and Development (OECD)** | - Recommendations from the council on artificial intelligence (OECD, 2023)<br>- OECD principles overview (OECD, 2024) |
| **United Nations** | - Interim Report: Governing AI for Humanity (UN, 2024) |
| **UNESCO** | - Ethics of Artificial Intelligence (UNESCO, 2024) |
| **World Health Organization (WHO)** | - Regulatory considerations on artificial intelligence for health (WHO, 2023)<br>- Ethics and governance of artificial intelligence for health, guidance on large multi-modal models (WHO, 2024) |
| **Regulatory Agencies** | |
| **EMA: European Medicines** | - Reflection paper on the use of Artificial Intelligence (AI) in the medicinal product lifecycle (EMA, 2023) |
| **Agency (European Union)** | |
| **FDA: US Food and Drug Administration (United-States)** | - AI and Medical products: How CBER, CDER and CDRH and OCP are working together (FDA, 2024)<br>- Using AI/ML in the development of drug and biological products (FDA, 2023)<br>- Artificial Intelligence in Drug Manufacturing (FDA, 2023)<br>- Predetermined Change Control Plans for Machine Learning-Enabled Medical Devices: Guiding Principles (FDA, 2023)<br>- AI/ML SAMD Action Plan (FDA, 2021)<br>- Good Machine Learning Practices for Medical Device Development: guiding principles (FDA, 2021) |
| **Medicines and Healthcare Products Regulatory Agency (MHRA) (UK)** | - Impact of AI on the regulation of medical products: Implementing the AI White Paper principles (MHRA, 2024) |
| **Multi-stakeholder groups** | |
| **Coalition for Health Artificial Intelligence (CHAI)** | - Blueprint for trustworthy AI implementation guidance and assurance for healthcare (CHAI, 2023) |
| **HTx H2020 (European Countries)** | - Recommendations to overcome barriers to the use of artificial intelligence-driven evidence in health technology assessment (Zemplenyi, 2023) |
| **NIST (United-States)** | - Artificial Intelligence Risk Management Framework (NIST, 2023)<br>- Towards a Standard for Identifying and Managing Bias in Artificial Intelligence (NIST, 2022) |
| **National Academy of Medicine (United-States)** | - AI in health, healthcare and biomedical science: an AI code of conduct (Adams et al. 2024) |

AI= Artificial Intelligence; AI/ML= Artificial Intelligence/Machine Learning; CBER= Center for Biologics Evaluation and Research; CDER = Center for Drug Evaluation and Research; CDRH= Center for Devices and Radiological Health; OCP= Office of Combination Products; SAMD= software as a medical device.